\documentclass{article}

\usepackage[final]{neurips_2020}

\usepackage[utf8]{inputenc} 
\usepackage[T1]{fontenc}    
\usepackage{url}            
\usepackage{booktabs}       
\usepackage{amsfonts}       
\usepackage{nicefrac}       
\usepackage{microtype}      
\usepackage{xcolor}         
\usepackage{enumitem}
\usepackage{graphicx}
\usepackage{multirow}
\usepackage{amsmath,amssymb} 
\usepackage[colorlinks,citecolor=blue,urlcolor=magenta,bookmarks=false,hypertexnames=true]{hyperref}
\usepackage{wrapfig}
\usepackage{algorithm}%
\usepackage{algorithmic}%
\usepackage{multirow}
\usepackage{multicol}
\newcommand*{\red}{\textcolor{black}}
\usepackage{natbib}

\title{Accelerating Deep Learning with Fixed Time Budget}
\author{%
  Muhammad Asif Khan\thanks{Corresponding author} \\
  Qatar Mobility Innovations Center (QMIC)\\
  Qatar University\\
  Doha, Qatar \\
  \texttt{mkhan@qu.edu.qa} \\
  \And
  Ridha Hamila \\
  Department of Electrical Engineering \\
  Qatar University\\
  Doha, Qatar \\
  \texttt{hamila@qu.edu.qa} \\
  \And
  Hamid Menouar \\
  Qatar Mobility Innovations Center (QMIC) \\
  Qatar University\\
  Doha, Qatar \\
  \texttt{hamidm@qmic.com} \\
}

\begin{document}
\maketitle
\begin{abstract}
The success of modern deep learning is attributed to two key elements: huge amounts of training data and large model sizes. Where a vast amount of data allows the model to learn more features, the large model architecture boosts the learning capability of the model. However, both these factors result in prolonged training time. In some practical applications such as edge-based learning and federated learning, limited-time budgets necessitate more efficient training methods. This paper proposes an effective technique for training arbitrary deep learning models within fixed time constraints utilizing sample importance and dynamic ranking. The proposed method is extensively evaluated in both classification and regression tasks in computer vision. The results consistently show clear gains achieved by the proposed method in improving the learning performance of various state-of-the-art deep learning models in both regression and classification tasks.
\end{abstract}

\section{Introduction}
Deep neural networks offer several benefits such as improved performance over challenging tasks using automatic feature extraction, capturing non-linear relationships in data, and transfer learning across multiple domains. However, these benefits come with challenges such as requiring huge computational resources for training deep learning models, long training time, and the need for large amounts of training data.
\par
In many practical situations, we often need to train a deep learning model on a large amount of training data with limited computational resources. Interestingly, the problem is exacerbated when there is a limited training time budget i.e., the training must be terminated after a specified time interval.
Although reducing the training time without compromising the model performance is generally desired, completing the training within a specified time frame becomes mandatory in some situations. Some examples of training with a fixed time budget are as follows. (i) In real-time prediction systems such as fraud detection and recommendation systems, the model needs to be periodically finetuned to keep up with the new data to make accurate predictions. (ii) In resource-constrained Internet of Things (IoT) systems \cite{DistributedIoT_Khan}, fixed training time is critical due to constrained energy consumption hardware resources. (iii) In rapid prototyping and model development, limited time is available to explore efficient model architectures and hyperparameters. (iv) In online learning, when new data is generated that might be out of distribution, models need to be updated in real time. (v) In a cloud-based environment, a limited financial budget allows short cloud service usage. In all such scenarios, efficient training strategies with a fixed time budget become significantly important.
\par
Over the years, several approaches have been proposed to reduce the training time of deep learning models.
First, when possible reducing the model size either by model pruning or replacing large models with shallow models can expedite the training. For instance, MobileNet \cite{MobileNetV2_2018} and SqueezeNet \cite{SqueezeNet_ICLR2017} are proposed for image classification, DroneNet \cite{DroneNet2023}, LCDnet \cite{LCDnet_khan2023} and MobileCount \cite{MobileCount_2019} are proposed for crowd density estimation, etc.
Second, by reducing the size of the training data, a technique called dataset pruning without compromising much on the accuracy can be effective when non-contributing data samples are removed from the training data. CLIP \cite{clip_khan2023} is an example of iterative dataset pruning to expedite training. A similar approach is a difficult sample resampling to eliminate the class imbalance to improve accuracy \cite{reviewer1a,reviewer1b}.
Third, using efficient training techniques such as curriculum learning \cite{curriculum_learning_ICML2009,CLCC_2024} without pruning the model parameters or eliminating training data samples, model training can be converged faster \cite{LCDnet_khan2023}. 
Fourth, using optimal training schedules e.g., learning rate schedule and learning rate decay functions \cite{li2019budgeted}, training can be expedited.
Though, all these methods can be effective in reducing training time to some extent, training a model within a fixed time budget while maximizing accuracy can be challenging \cite{TCL_2023}. Possible reasons include the unavailability of high-quality data, limited computing resources, search space complexity, hyper-parameter tuning, and trade-offs between model complexity, training duration, and computational resources.
Our goal in this work is to improve the model learning performance under a fixed training time budget using the data-centric approach. Several works recently have shown the effectiveness of dataset pruning \cite{Yang2022,Paul2021,Katharopoulos2018} in improving model learning and convergence. Dataset pruning generally involves the selection of a subset or elimination of samples from the original dataset to reduce the effective size of the dataset used in the training of a model. Apart from this, data sampling is also used for outlier and noise removal, class balancing, and bias reduction. The technique has also been used to improve learning performance in curriculum learning (CL) \cite{clip_khan2023}. CL is a training method in which the training data is arranged (rather than random) and presented to the model over time rather than all at once \cite{curriculum_learning_ICML2009,Guy_2019,Guo_2018,Li_2021}. 
\par
The selection of subsets from large training sets follows the notion that not all training samples essentially contribute to the training process, and one can eliminate these non-contributing samples from the training data \cite{Katharopoulos2018,clip_khan2023}. However, the criteria for defining sample contribution (importance), ranking the importance, and approaches for the subset selection differ. Two methods to compute sample importance are gradient norm and loss. As computing gradient norms can be significantly expensive to compute during the training \cite{Katharopoulos2018}, we use loss-based importance scoring.
\par

The contributions of this work are summarized as follows:
\begin{itemize}
\item We propose a data selection strategy to iteratively select a mix of more important/informative samples and representative samples from the full dataset to dynamically acquire a more representative subset to train a deep learning model. This achieves a balance between learning and fairness enabling faster model convergence.
\item We propose an algorithm to dynamically set important parameters for data sampling and define the maximum number of training iterations to ensure the training completes within the time budget with maximum achievable performance.
\item We evaluate the performance of the proposed algorithm and data sampling strategy in two distinct tasks in computer vision i.e., image classification and crowd density estimation (regression). We performed extensive experiments using 6 models over 2 datasets for image classification and 7 models over 2 datasets for crowd density regression.
\end{itemize}

\section{Related Work}


Authors in \cite{clip_khan2023} used the contribution of individual examples (training samples) to remove non-contributing training samples from the original training set and applied curriculum learning to reduce the total training time. Dataset pruning has also been investigated in \cite{Yang2022}, \cite{Paul2021}, and \cite{Zayed2022}.

In \cite{Yang2022}, authors removed redundant training samples based on the loss function. More specifically, it distinguishes and removes redundant training examples based on the loss function. Their results report that the accuracy of the model utilizing the pruned dataset (with a lower number of samples than the original dataset) is comparable (slightly lower) to the model trained on the complete dataset.
In \cite{Paul2021}, authors used gradient normed and L2-norm score functions to calculate the feature importance and used the two score functions to remove samples during the earlier training stage yielding a compact dataset after a few iterations. The authors report up to 50\% accuracy gain using the compact dataset versus the full dataset.
In \cite{Zayed2022}, authors use dataset pruning as a superior alternative method to data augmentation in reducing bias in the data and improving the fairness of the model.
Similarly, authors in \cite{Katharopoulos2018} used per-importance scores to improve the learning performance in stochastic gradient descent (SDG) by reducing the variance and proposed a method to use importance-based sampling automatically when it is beneficial during the entire training phase.
\par
Unlike the works \cite{Yang2022,Paul2021,Zayed2022,clip_khan2023}, authors in \cite{Covert2020} studies the impact of individual feature importance on model convergence. In \cite{Covert2020}, a model-agnostic method called SAGE is proposed for calculating the global feature importance. SAGE determines a model’s dependence on each feature by applying the Shapley value to a function representing the predictive power contained in subsets of features. Similarly, authors in \cite{Casalicchio2018}, study the impact of changes in the features impact the model performance using local feature importance using partial importance (PI) and individual conditional importance (ICI) plots. The authors also proposed Shapely feature importance as a tool to compare feature importance across different models. Several methods have been proposed to calculate feature importance e.g., partial dependence (PD) plots \cite{Friedman2001}, individual conditional expectation (ICE) plots \cite{Goldstein2013}, and SHAP values \cite{Lundberg2018}. However, these methods may adversely increase the complexity and can be computationally expensive for resource-limited devices.

The aforementioned works all put emphasis on the quality of data and propose various techniques to refine the training data and sampling training data to improve the model performance and efficiency. However, these works focus on training a model on a compact dataset to achieve comparable performance. These studies do not consider the time budget and it may take longer to achieve comparable accuracy using the compact dataset due to permanently removing a significant portion of data. We found these works \cite{TCL_2023,li2019budgeted,Katharopoulos2018} that focus on the limited resource budget. However, \cite{li2019budgeted} studies the impact of learning rate schedules to reduce the training time and does not consider a fixed training time budget. \cite{TCL_2023} considers a fixed time budget but it considers a teacher-student model to train a student network. Similarly, \cite{Katharopoulos2018} considers a fixed time budget but it works for SGD only and may not be effective in minibatch training. We believe these methods are certainly useful but do not fit the purpose of training a target model without requiring a teacher model (unlike \cite{TCL_2023}), without fine-tuning to find optimal learning rate schedules (unlike \cite{li2019budgeted}) and in standard mini-batch training set up (unlike \cite{Katharopoulos2018}).

\section{The Proposed Method}

In this section, we present the proposed method to train deep learning models with a fixed training time budget. The time budget may come from the application or the computing resources.
Consider a model denoted as $M$ trained on a full training dataset $X$ and the restricted model is denoted as $\mathcal{M}$ when trained on a subset of the original dataset $X_s$. We are interested in dynamically selecting $X_s$ such that a model $\mathcal{M}$ trained on $X_s$ achieves either (i) better performance than $M$ when both are trained for an equal amount of time $T$, or (ii) comparable performance as $M$ in less time than $T$ where $T$ is the training time of model $M$.




We consider some essential principles of machine learning to design our strategy.
When selecting a smaller subset $X_s$, we can intuitively reduce the training time for a given number of training iterations. However, permanently eliminating a large number of samples would also affect the accuracy and is thus not always a good strategy.
Thus designing a selection strategy that dynamically chooses samples, we can avoid the loss of significant information to improve the learning performance. 
Secondly, all training samples contribute differently. Hence, selecting easy samples in the $X_s$ would allow $\mathcal{M}$ to converge quickly but it may fail to generalize and thus perform poorly on unseen data. On the other hand, difficult samples typically contain more information and are generally considered more helpful to improve the model accuracy \cite{TCL_2023}, but would slow down convergence as it takes more time to learn the features in such samples. Thus, a good trade-off would be to consider a mix of both easy and difficult samples for inclusion in the subset $X_s$ to achieve both speed and accuracy.
Third, when choosing the subset $X_s$, a good criterion to compute feature importance and ranking is vital for both learning and speed.
\par
\begin{figure}[htbp]
    \centering
    \includegraphics[width=0.6\columnwidth]{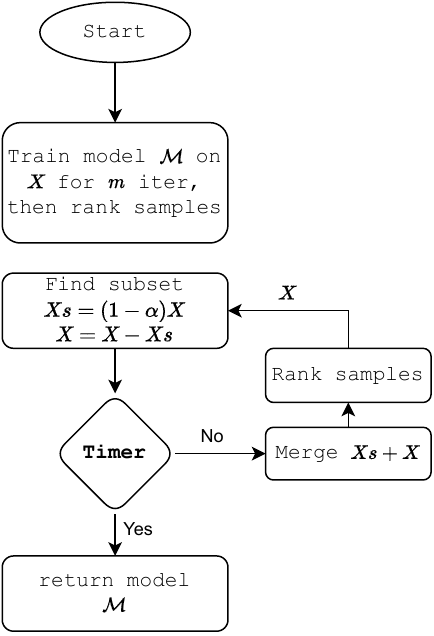}
    \caption{A simple illustration of the proposed dynamic training strategy with selective curriculum and sample importance.}
    \label{fig:prop_scheme}
\end{figure}


\begin{algorithm}[ht]
\caption{\textbf{\textit{TFTB: Train with Fixed Time Budget}}}
\label{alg:algorithm}
\textbf{Input}: full dataset $X$ with $B$ mini-batches, scoring function $f$, time budget $T$, iterations $m$, sampling ratio $\alpha$\\
\textbf{Output}: trained model $\mathcal{M}$

\begin{algorithmic}[1] 
\STATE Record remaining time $T^*$
\STATE Train $\mathcal{M}$ using $X$ = $\frac{X}{bs}$ mini-batches. for a few (e.g., $m$) iterations.
\STATE $\mathcal{M} \left(X\right)$. \\
\STATE Compute batch processing time. \\
\STATE $tb=(current\_time - start\_time)/(B*m)$.
\STATE Update time budget $T=T-(current\_time - start\_time)$

\STATE Compute per-sample importance ($I_i$) for $i$ samples.
\STATE Rank samples as per $I_i$.
\STATE $ I_{\sigma(1)} \geq I_{\sigma(2)} \geq I_{\sigma(3)} \geq \cdot \geq I_{\sigma(X)} $
\STATE Calculate subset $X_s$:
\STATE $X_s = X * (1- \alpha)$
\STATE Update $X$:
\STATE $X = X - X_s$
\STATE Calculate remaining iterations for $X_s$.
\STATE $num\_iterations = \frac{X_s}{bs} * tb$
\FOR {iteration in range(0, num\_iterations)}
    \STATE $\mathcal{M} \left( X_s \right)$. \\
    \STATE Compute $I_i$.
    \STATE Merge: $Xs + X$
    \STATE Rank samples in $X$ as per $I_i$.
    \STATE Select subset for next iteration.
    \STATE $X_s = X * (1- \alpha)$.
\ENDFOR
\STATE \textbf{return} $\mathcal{M}$
\end{algorithmic}
\end{algorithm}
Based on the aforementioned principles, we designed a dynamic strategy for computing sample importance scores, ranking the samples based on their importance scores, and iteratively updating the subset $X_s$ during the training phase. We further design an algorithm called \textit{TFTB: Train with Fixed Time Budget)} that incorporates the dynamic sampling strategy and ensures that the training completes within the available fixed time budget and achieves the maximum accuracy. The algorithm works as follows:
\par
The model $\mathcal{M}$ with a fixed time budget $T$ is first trained on a full dataset $X$ for $m$ iterations. The time taken to run these initial iterations is subtracted from $T$ to update the remaining time. These initial iterations on $X$ serve three purposes. (i) The model initially learns from the random samples and the gradient flow starts more naturally. (ii) The batch processing time is calculated from these iterations which gives an estimation of the remaining iterations that are likely to be completed within the remaining time budget $T$. (iii) The per-sample losses which serve as the importance scores ($I_i$) are calculated from these iterations to initially rank samples as follows:

$$ I_{\sigma(1)} \geq I_{\sigma(2)} \geq I_{\sigma(3)} \geq \cdot \geq I_{\sigma(X)} $$

where $I_{\sigma(i)}$ represents the importance score of the $i^{th}$ ranked sample and $\sigma_{(i)}$ is the permutation of indices that represents the rank of the $i^{th}$ sample based on its importance score.

Once the samples are ranked based on their individual importance score, the subset $X_s$ is calculated by selecting top-ranked samples as follows:

$$ X_s = (1-\alpha) X$$
where $\alpha$ is the sampling ratio i.e., $\alpha=0.1$ means that only 10\% samples are excluded from $X_s$, $\alpha=0$ means no reduction or full dataset ($X$).
Once the subset $X_s$ is computed, the remaining training iterations always use the reduced dataset $X_s$, whereas the dataset $X$ is reduced to contain only the eliminated samples ($X=X - X_s$). However, at the end of each training iteration (which can also be set to more than one iteration), the per-sample importance scores are recalculated. Using the new scores, the $X_s$ and $X$ are merged and the samples are ranked again to select subset $X_s$ for use in the subsequent training iteration. The importance scores are weighted using the variance of score values. The process is repeated for the remaining training iterations until the iterations reach the maximum number of iterations (calculated earlier) or the time budget is fully utilized.
The process of model training using the proposed method is illustrated in Fig. \ref{fig:prop_scheme}.
To ensure that the training using the proposed sampling strategy is completed within the available time budget $T$, the process is formalized in Algorithm \ref{alg:algorithm}.
\par

\subsection{Theoretical Foundation}
\red{
The adaptive sampling strategy is based on the idea that different samples contribute differently to the model’s learning process. In our method, the importance scores are computed based on the per-sample losses during training. The intuition here is that samples with higher losses indicate that the model is struggling to learn those specific examples, suggesting that they contain more valuable information. Conversely, samples with lower losses are easier for the model to learn and thus provide diminishing returns during training.
To dynamically adjust the subset $X_s$, we periodically recalculate these importance scores as training progresses. This allows us to adapt the dataset in real-time by including or excluding samples based on their evolving contribution to the learning process.
At each iteration (or after a set of iterations), the importance scores are recalculated and ranked. The variance of the scores is used to weight their influence in selecting the subset $X_s$. Samples with a higher variance in their importance scores are more likely to be retained, reflecting their higher learning potential across iterations. The sampling ratio $\alpha$ is a key parameter that controls how aggressively we reduce the dataset, allowing for a mix of easy and difficult samples to be retained.
The theoretical foundation of this method is rooted in two well-established concepts: curriculum learning and importance sampling. Curriculum learning suggests that training a model with a gradual introduction of more difficult examples improves generalization. Importance sampling, on the other hand, prioritizes samples based on their significance to the model’s learning. Our method combines these principles by dynamically selecting samples that balance ease of learning and information content.}

\subsection{Balancing Sample Diversity and Importance}
\red{
The sample importance scores are dynamically adjusted based on the per-sample loss values computed at each training iteration (or a set of iterations). Initially, the model is trained on the full dataset $X$ for a few iterations. During these initial iterations, the loss for each sample is tracked and serves as the first estimation of its importance. Higher loss values indicate that a sample is difficult for the model to learn, suggesting that it carries more useful information. As training progresses, the importance scores are recalculated after each iteration (or after a predefined number of iterations) using the updated per-sample losses. The scores are then used to rank the samples, enabling the algorithm to dynamically adjust the subset $X_s$ used for the subsequent training iterations.
A key challenge is ensuring that the subset $X_s$ contains both high-importance samples (which are typically more difficult) and a sufficient level of diversity (which may include easier samples that ensure better generalization). We address this trade-off in two ways:
}
\begin{enumerate}
    \item \red{The importance scores are not directly used for selection; they are weighted by their variance across iterations. This weighting mechanism helps to balance the inclusion of diverse samples, as those with high variance indicate they are impactful in specific stages of training, while those with consistently high scores are prioritized for learning.}
    \item \red{The sampling ratio  $\alpha$ controls how aggressively the dataset is reduced. A lower $\alpha$ retains more samples, ensuring diversity, while a higher $\alpha$ focuses on a more refined subset of high-importance samples. This ratio is dynamically adjusted based on the model’s convergence behavior: if the model is converging too slowly, $\alpha$ is reduced to include more diverse samples to avoid overfitting; if convergence is fast, $\alpha$ is increased to focus on high-impact samples.}
\end{enumerate}

\red{
The strategy also uses curriculum learning, where easy-to-learn samples are initially prioritized to help the model stabilize and learn basic features. As training progresses, the strategy shifts to include more difficult samples that are critical for refining performance.
}

\subsection{Computational Overhead}
\red{
The dynamic ranking and importance scoring in TFTB introduce additional computational overhead that comes from sample importance score calculation and iterative sample ranking. While these steps introduce extra computational demands, TFTB is designed to offset this overhead by reducing the overall number of samples processed during each iteration. This reduction in training data can more than compensate for the added overhead, leading to net time savings. Moreover, The extra computation overhead of TFTB can be reduced by using techniques such as (i) batch-loss or stochastic estimated loss instead of actual per-sample loss, and (ii) efficient ranking methods such as top-K selection algorithms and sparse updates (update only samples that show significant change).
The empirical analysis in this study show that as the dataset size increases, the time savings from reduced iterations dominate the additional overhead introduced by TFTB, making it more efficient for larger datasets.
}
It is worthy to note that that the proposed method samples a subset from the full dataset in every iteration, it is typically recommended to use it on datasets of reasonable size. However, the noisy data should not impact the performance when compared to standard training \cite{ref1,ref2}.

\section{Experiments and Results}

\subsection{Learning Tasks}
We evaluate the performance of the proposed method in two learning problems in computer vision i.e., image classification and regression. For the regression task, we consider the use case of crowd density estimation. Crowd density estimation is a widely used technique for crowd counting which is an interesting problem in computer vision \cite{Khan2022RevisitingCC,Gao2020CNNbasedDE,Fan2021ASO,Yang2022SurveyOA,Bai2020ASO}.

\subsection{Datasets}
We used the CIFAR-10 and CIFAR100 datasets. Due to the small size of the datasets, these are widely used in several investigative studies for finding novel model architectures and training techniques \cite{li2019budgeted,Guy_2019}.

\paragraph{CIFAR10:} The CIFAR10 dataset consists of 60000 color images each of size 32x32. The images are divided into 10 classes, with 6000 images per class. The dataset has been divided into a training set of 50000 images and a test set of 10000 images. The image classes are mutually exclusive (no overlapping between classes).

\paragraph{CIFAR100:} The CIFAR100 dataset has 100 classes containing 600 images each. There are 500 training images and 100 testing images per class.
\par
For crowd counting task, we used two datasets of human crowds (ShanghaiTech Part-B \cite{MCNN_CVPR2016}) and vehicle crowds (CARPK \cite{CARPK_dataset}).

\paragraph{ShanghaiTech Part-B:} The dataset is a large-scale crowd-counting dataset used in many studies. The dataset is split into train and test subsets consisting of 400 and 316 images, respectively. All images are of fixed size $(1024 \times 768)$.

\paragraph{CARPK:} This dataset contains images of cars from 4 different parking lots captured using a drone (Phantom 3 Professional) at approximately 40-meter altitude. The dataset contains 1448 images split into train and test sets of sizes 989 and 459 images, respectively. The dataset contains a total of 90,000 car annotations and has been used in several object counting and object detection studies.
\par
\paragraph{Density Maps Generation:}
In both crowd-counting datasets, the images are point annotated The annotation comes in the form of a binary matrix (of the same size as the image) of pixel values such that a value of 1 denotes the head position whereas a 0 represents no head. The binary matrix which is called a dot map or localization map is used to create density maps that serve as the ground truth for the images to train the model. A density map is generated by convolving a delta function $\delta(x - x_i)$ with a Gaussian kernel $G_\sigma$, where $x_i$ are pixel values containing the head positions.

\begin{equation}
    D = \sum_{i=1}^{N}{ \delta(x-x_i) * G_\sigma}
\end{equation}

where, $N$ denotes the total number of dot points with value 1 in the dot map (i.e., total headcount in the input image). The loss function is the $l_2$ distance (euclidean distance) between the target and the predicted density maps (Eq. \ref{eq:mse_loss}).

\begin{equation} \label{eq:mse_loss}
    L(\Theta) = \frac{1}{N} \sum_{1}^{N}{ ||D(X_i;\Theta) - D_i^{gt}||_2^2}
\end{equation}

where $N$ is the total number of samples in training data, $X_i$ is the input image, $D_i^{gt}$ is the ground truth density map, and $D(X_i;\Theta)$ is the predicted density map. 



\subsection{Evaluation Metrics}
We used mean accuracy as the commonly used metric for classification tasks which is defined as the ratio of correct predictions to the total number of predictions for all classes.
\par
For crowd counting, we used the two standard and widely used metrics i.e., mean absolute error (MAE) and mean squared error (MSE) which are calculated using the following Eq. \ref{eq:mae} and  \ref{eq:mse}.

\begin{equation} \label{eq:mae}
    MAE = \frac{1}{N} \sum_{1}^{N}{\|e_n - \hat{g_n}\|}
\end{equation}

\begin{equation} \label{eq:mse}
    MSE = \frac{1}{N} \sum_{1}^{N}{(e_n - \hat{g_n})^2}
\end{equation}

where, $N$ is the size of the dataset, $g_n$ is the target or label (actual count) and ${e_n}$ is the prediction (estimated count) in the $n^{th}$ crowd image.

\subsection{Baselines}
We trained three well-known off-the-shelf classification models i.e., GoogleNet \cite{GoogLeNet_2015}, ResNet18 \cite{ResNet_2016}, MobileNetv2 \cite{MobileNetV2_2018} and three custom-defined CNN classification models of varying sizes (small, medium, and large).
The small model consists of two convolution (Conv) layers and three fully connected (FC) layers. The first Conv layer has 6 $(5\times5)$ filters whereas the second Conv layer has 16 $(5\times5)$ filters.
The medium model consists of three Conv layers each containing 32, 64, and 128 filters of size $(3\times3)$, respectively, followed by a FC layer.
The large model has six Conv layers all containing $3\times3$ filters. The first two layers contain 32 layers, the third and fourth layers contain 64 filters, whereas the last two layers contain 128 filters. There is one FC layer at the end.
All custom models use ReLU activation and a $2\times2$ Max Pool layer after each Conv layer.
\par

For crowd density estimation, we trained seven off-the-shelf crowd counting models i.e., MCNN \cite{MCNN_CVPR2016}, CMTL \cite{CMTL_AVSS2017}, CSRNet \cite{CSRNet_CVPR2018}, TEDnet \cite{TEDnet_CVPR2019}, ASNet \cite{ASNet_CVPR2020}, SASNet \cite{SASNet_AAAI2021}, and DroneNet \cite{DroneNet2023}. These models are carefully selected due to their unique architecture to make our evaluation more fair (independent of model architectures). For instance, MCNN is a simple multi-column network, CMTL is a cascaded architecture of two CNN networks, CSRNet is a two-stage architecture with a VGG16 frontend and a backend with dilated convolution, TEDnet follows an encoder-decoder structure, ASNet has two networks with attention scaling, SASNet is a pyramid network with a different loss function, and DroneNet is an MCNN alternative using non-linear operations instead of convolution.
\subsection{Settings}
In the first set of experiments on image classification, we trained the six baseline models on CIFAR10 and CIFAR100 datasets respectively. For reproducibility, we kept most of the parameter settings constant. For instance, we used \textit{Adam} optimizer with a fixed learning rate $1\times10^{-2}$ without a learning rate scheduler. We did not use any image augmentations. All the models are trained for 20 epochs (Note: One epoch in TFTB is derived such that the model is exposed to the number of samples equal to the standard dataset size.). To train the models using our proposed method (Algorithm \ref{alg:algorithm}), we used two different values of $\alpha={0.3, 0.4}$ which means that the subset $X_s$ eliminates $30\%$ and $40\%$ of the samples of $X$ in each set of experiments. We used per-class sampling with an equal value of $\alpha$ to avoid class imbalance.
Though we used separate train and test sets to evaluate the performance of our methods on absolutely unseen samples, we explicitly used $early\_stopping$ with a patience value of 5 to avoid overfitting i.e., when the validation loss is not reducing for five epochs, irrespective of the training loss is reducing, the training would halt. This is necessary because we did not use augmentation in our experiments.
\begin{table*}[ht]
\caption{Summary of experimental settings.}
\label{tab_settings}
\centering
\resizebox{.98\columnwidth}{!}{
\setlength{\tabcolsep}{6pt}
\renewcommand{\arraystretch}{1}
\begin{tabular}{|p{3cm}|p{5.5cm}p{5.5cm}|} \toprule
& \textbf{Classification} & \textbf{Regression} \\ \midrule
&& \\[-0.5em]       
Tasks  & Image classification  & Density estimation \\[0.5em]
Datasets & CIFAR10, CIFAR100  & ShanghaiTech-PB, CARPK \\[1em]
Baselines & VGG16, ResNet18, MobileNet\_V2, Custom Nets (small, medium, large)  &MCNN, CMTL, CSRNet, TEDnet, ASNet, SASNet, and DroneNet. \\ [2em]
Metrics & Accuracy & MAE, MSE \\ [0.5em]
Sampling ratio ($\alpha$) & \{0.3, 0.4\} & \{0.2, 0.3\}\\ [1.5em]
Optimizer & Adam & Adam \\ [0.2em] \bottomrule
\end{tabular}
}
\end{table*}
In the second set of experiments on crowd density estimation, we used a fixed value of $\sigma$ for all images in the same dataset to generate the ground truth density maps i.e., $15$ and $10$ for ShanghaiTech Part B and CARPK datasets, respectively.
For training the crowd models, we used original image sizes for training and a batch size of 32. We did not use any image augmentation (for reproducibility purposes). We used the $Adam$ optimizer with a learning rate of $1 \times 10^{-3}$ and zero weight decay. We used pixel-wise $L_2$ loss (i.e., mean squared error (MSE)) function. All the baseline models are trained for 30 epochs. Due to the small number of images in these datasets and large variations in crowd densities across different images, we kept the value of $\alpha=0.2$ (i.e., eliminated $20\%$ of the total samples from $X$) for ShanghaiTech Part-B and $\alpha=0.3$ for CARPK dataset. We used the PyTorch framework running on a Lambda machine with two RTX-8000 GPUs to train all models.
Table \ref{tab_settings} provides a summary of experimental setups used in this paper.
\subsection{Results}
In this section, we present the results of experiments and compare the performance of the proposed method (\textit{TFTB}). The results are discussed separately for image classification and regression (density estimation) tasks.


\begin{figure*}[ht]
    \centering
    \includegraphics[width=0.9\textwidth]{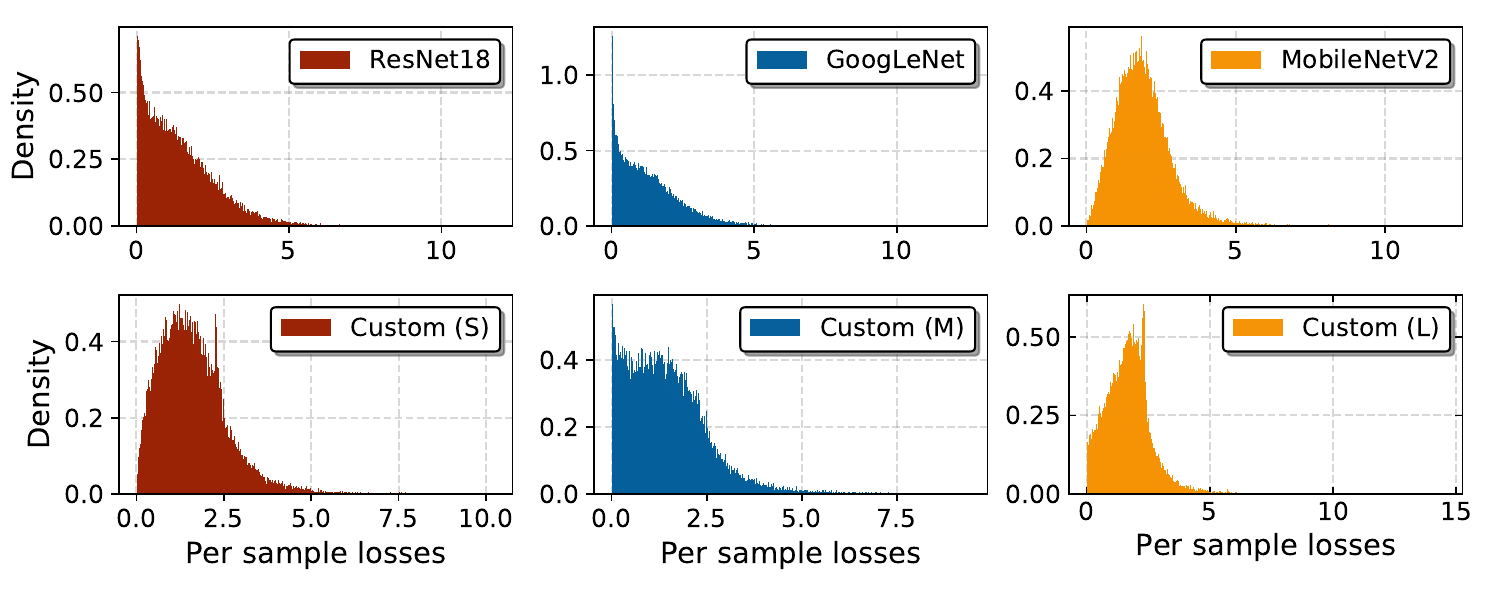}
    \caption{Per-sample loss values for the CIFAR10 dataset computed during the initial training phase, showing large variations in loss values for different baseline models.}
    \label{fig:per_sample_losses}
\end{figure*}

\begin{figure*}[!h]
    \centering
    \includegraphics[width=0.9\textwidth]{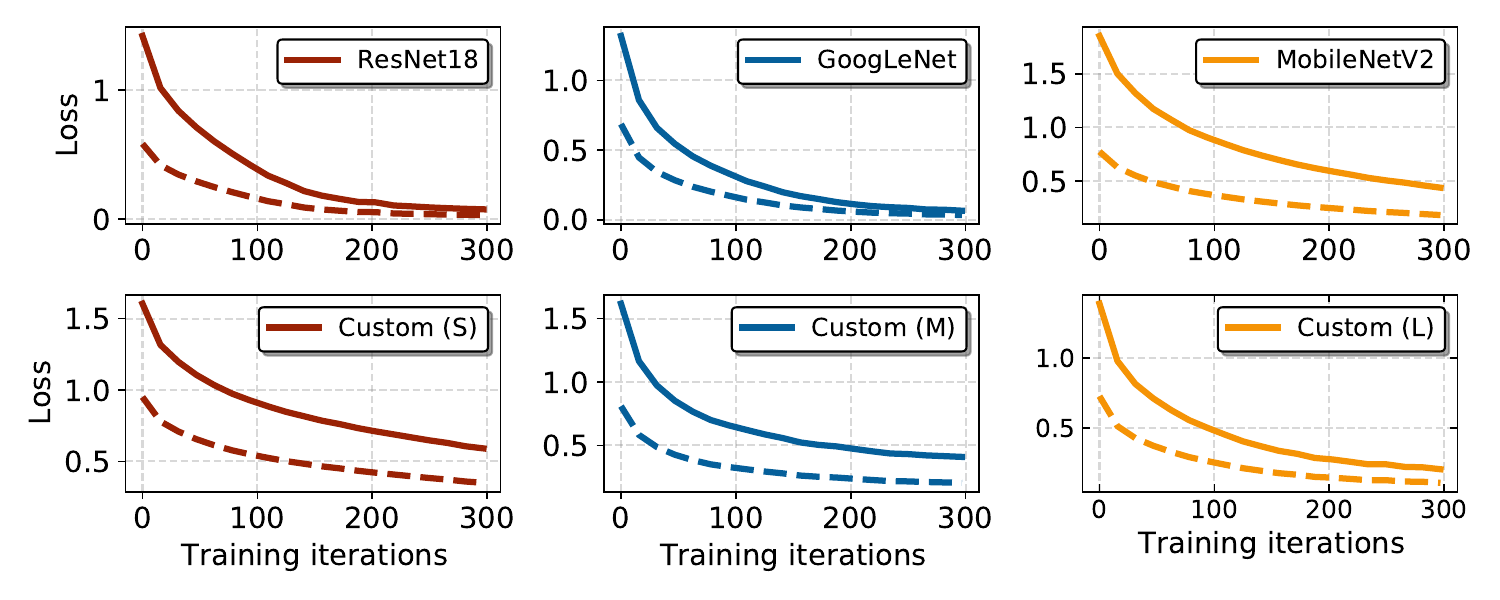}
    \caption{Comparison of loss convergence in image classification using the baseline model with standard training (solid lines) on the full dataset (CIFAR10) versus selective samples using the proposed method (dashed lines).}
    \label{fig:loss_curves}
\end{figure*}

    

\subsubsection{Performance in Classification}
In image classification, the six baseline models are trained first using standard training on full datasets, and the accuracy is recorded. Then the models are trained from scratch using the \textit{TFTB} method. In \textit{TFTB}, each model is initially trained for a few iterations using random sampling from full datasets to compute per-sample losses and rank samples based on the importance scores. The distribution of per-sample losses for all six models for the CIFAR10 dataset is presented in Fig. \ref{fig:per_sample_losses}. It can be visualized that each model has a different learning characteristic and thus different losses are calculated for the same dataset. This analysis provides an insight that although the \textit{TFTB} in general is model-agnostic, the importance scores calculated in \textit{TFTB} are model-specific and may not benefit other models to the same extent. This is the reason why \textit{TFTB} incorporates the dynamic sample ranking inside the training loop.
\par

\begin{figure}[!h]
    \centering
    \includegraphics[width=0.8\columnwidth]{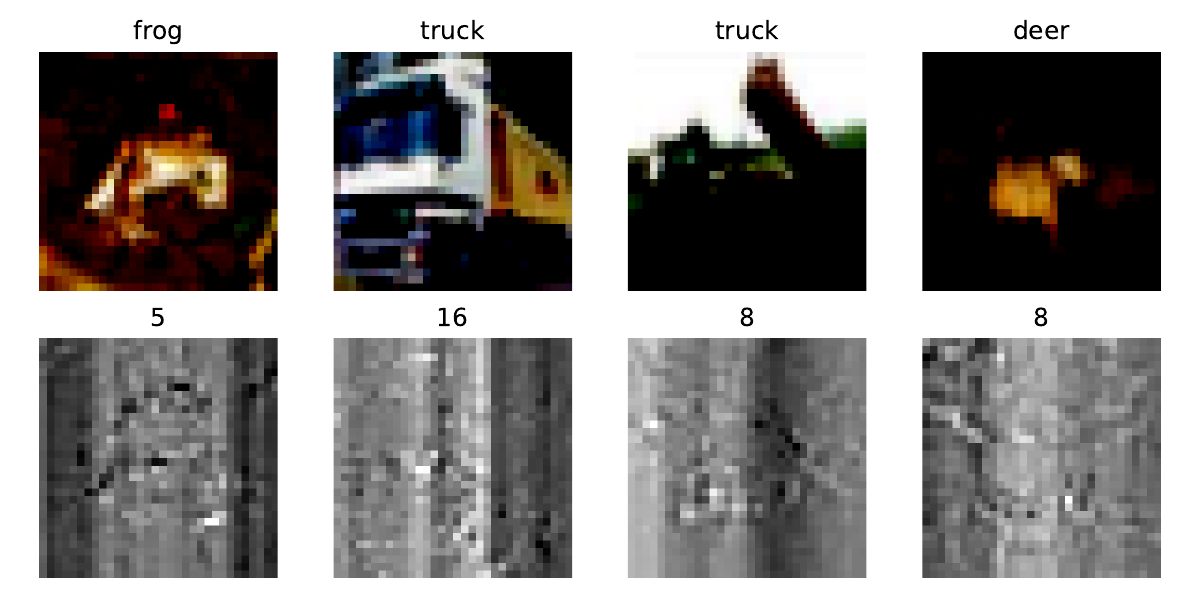}
    \\
    \includegraphics[width=0.8\columnwidth]{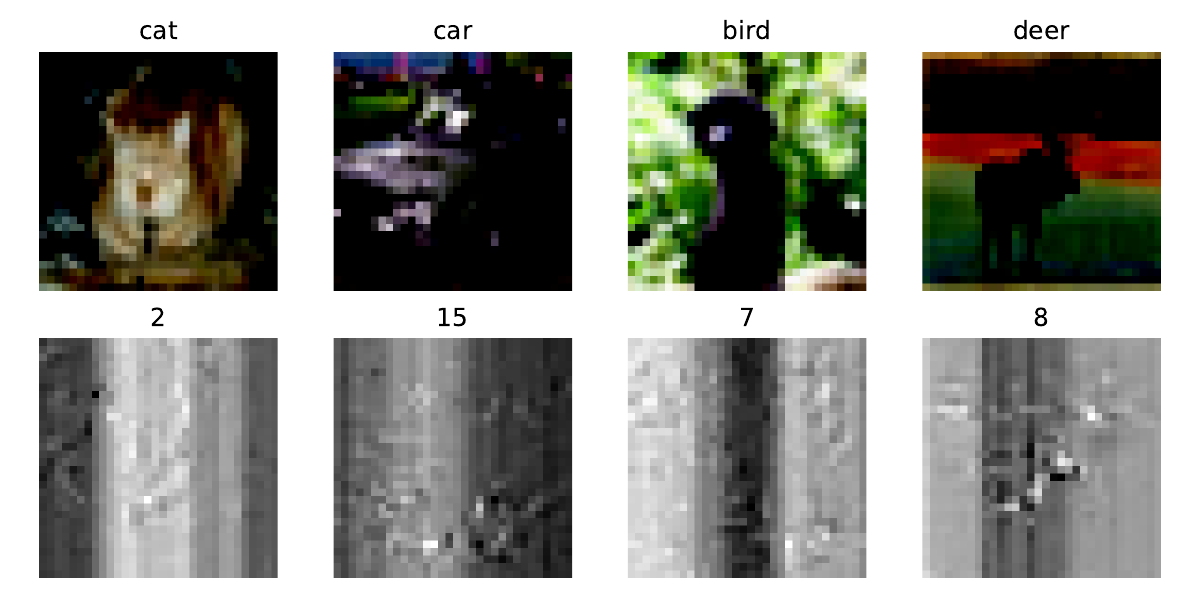}
    \caption{Illustration of important features using integrated gradients method for various image classes in CIFAR10 dataset (zoom in required for better visualization of features).}
    \label{fig:igradients}
\end{figure}

Fig. \ref{fig:loss_curves} shows the loss curves for all models when trained on the CIFAR10 dataset and compares the loss reduction using \textit{TFTB} against the standard training. For fairness, the loss values are plotted against training iterations where in each iteration the model is exposed to the same number of samples in both methods with the fact that standard training uses random sampling whereas the proposed method uses samples selected by the \textit{TFTB} method (Algorithm \ref{alg:algorithm}). To further investigate the learning capabilities of the trained models using the \textit{TFTB} method, Fig. \ref{fig:igradients} presents the important features for randomly selected individual samples from CIFAR10 dataset using integrated gradients method.
The faster loss convergence in the \textit{TFTB} method in Fig. \ref{fig:loss_curves} indicates better learning performance. The model generalization performance on unseen text data is then tested for both CIFAR10 and CIFAR100 datasets. The results for CIFAR10 and CIFAR100 are presented in Table. \ref{tab_cifar10} and Table. \ref{tab_cifar100}, respectively.
\par




\begin{table}[ht]
\centering
\caption{Accuracy comparison of several classification models over CIFAR10 dataset.}
\setlength{\tabcolsep}{8pt}
\renewcommand{\arraystretch}{1}



\begin{tabular}{|r|c p{1.2cm} p{1.3cm}|} \hline
\textbf{Model} & \textbf{Baseline} &\textbf{Ours $\alpha=0.3$} &\textbf{Ours $\alpha=0.4$}  \\ \toprule 
ResNet18 &76.0 &\textbf{81.2}    &77.2  \\[0.2em] 
GoogLeNet &86.2 &\textbf{89.5}    &88.6  \\[0.2em] 
MobileNet\_V2 &75.3 &\textbf{79.2}  &77.3  \\[0.2em] 
Custom (small) &63.5 &\textbf{66.7}  &64.2  \\[0.2em] 
Custom (medium) &75.5 &\textbf{77.9}  &77.3  \\[0.2em] 
Custom (large) &76.4 &\textbf{79.5}    &76.1 \\[0.2em] \bottomrule
\end{tabular}

\label{tab_cifar10}
\end{table}




\begin{table}[ht]
\centering
\caption{Accuracy comparison of several classification models over CIFAR100 dataset.}
\setlength{\tabcolsep}{8pt}
\renewcommand{\arraystretch}{1}


\begin{tabular}{|r|c p{1.2cm} p{1.3cm}|} \hline
\textbf{Model} & \textbf{Baseline} &\textbf{Ours $\alpha=0.3$} &\textbf{Ours $\alpha=0.4$}  \\ \toprule 
ResNet18 &46.2 &\textbf{51.7}   &50.4 \\[0.2em] 
GoogLeNet &63.6 &\textbf{67.8}  &67.1  \\[0.2em] 
MobileNet\_V2 &35.4 &\textbf{43.4} &40.7  \\[0.2em] 
Custom (small) &30.2 &\textbf{32.6} &31.1   \\[0.2em] 
Custom (medium) &43.1 &\textbf{47.2} &42.7   \\[0.2em] 
Custom (large) &45.2 &\textbf{46.7}  &44.9 \\[0.2em] \bottomrule
\end{tabular}
\label{tab_cifar100}
\end{table}

Our results consistently show that the proposed method (\textit{TFTB}) outperforms all baselines on both CIFAR10 and CIFAR100 datasets. The accuracy is generally higher for $\alpha=0.3$, but even better results are achieved for $\alpha=0.4$ than standard training in most cases. The consistency in the performance gains over all baselines on both datasets shows the robustness of the proposed method in the image classification task.
In the next section, we provide the results of our ablation study to evaluate the performance of \textit{TFTB} in the crowd density regression task.
\vspace{0em}

\begin{table*}[hbt]
\centering
\caption{Comparison of Mean Absolute Error (MAE) and Mean Squared Error (MSE) over ShanghaiTech Part B and CARPK datasets.}
\setlength{\tabcolsep}{6pt}





\begin{tabular}{|r|r| cccc |cccc|} \toprule[0.05em]
\multicolumn{1}{|c|}{\multirow{3}{*}{\textbf{Model}}} & \multicolumn{1}{c|}{\multirow{3}{*}{\textbf{Venue}}} & \multicolumn{4}{c|}{\textbf{ShanghaiTech Part B}}  & \multicolumn{4}{c|}{\textbf{CARPK}}                                        \\ \cmidrule{3-10} 

\multicolumn{1}{|c|}{}  & \multicolumn{1}{c|}{} & \multicolumn{2}{c|}{\textbf{Baseline}}   & \multicolumn{2}{c|}{\textbf{Ours}}   & \multicolumn{2}{c|}{\textbf{Baseline}}  & \multicolumn{2}{c|}{\textbf{Ours}} \\[0.1em] \cmidrule{3-10} 

& &MSE &MAE &MSE &MAE &MSE &MAE &MSE &MAE \\[0.1em]  \midrule[0.1em]                   
MCNN   &CVPR2016  &69.8 &40.2 &62.5 &\textbf{37.1}    &38.5 &19.2 &29.5 &\textbf{15.4} \\[0.4em]
CMTL   &AVSS2017  &79.6 &36.2 &73.8 &\textbf{34.8}    &36.7 &17.3 &32.1 &\textbf{14.6} \\[0.4em]
CSRNet &CVPR2018  &46.5 &28.7 &42.3 &\textbf{24.4}    &26.2 &12.5 &21.5 &\textbf{10.2} \\[0.4em]
TEDnet &CVPR2019  &74.2 &35.2 &65.6 &\textbf{33.8}    &32.8 &14.7 &28.3 &\textbf{11.4} \\[0.4em]
ASNet  &CVPR2020  &72.8 &34.0 &67.7 &\textbf{31.5}    &29.5 &12.3 &23.7 &\textbf{10.8} \\[0.4em]
SASNet &AAAI2021  &58.3 &30.2 &52.4 &\textbf{28.5}    &31.3 &14.6 &25.2 &\textbf{10.1} \\[0.4em]
DroneNet &CCNC2023 &65.4 &38.1 &59.8 &\textbf{36.2}   &34.6 &18.5 &30.9 &\textbf{15.0} \\[0.4em] \bottomrule[0.1em]
\end{tabular}

\label{tab_counting}
\end{table*}

\subsubsection{Performance in Regression (Density Estimation)}
\begin{figure}[!h]
    \centering
    \includegraphics[width=0.85\columnwidth]{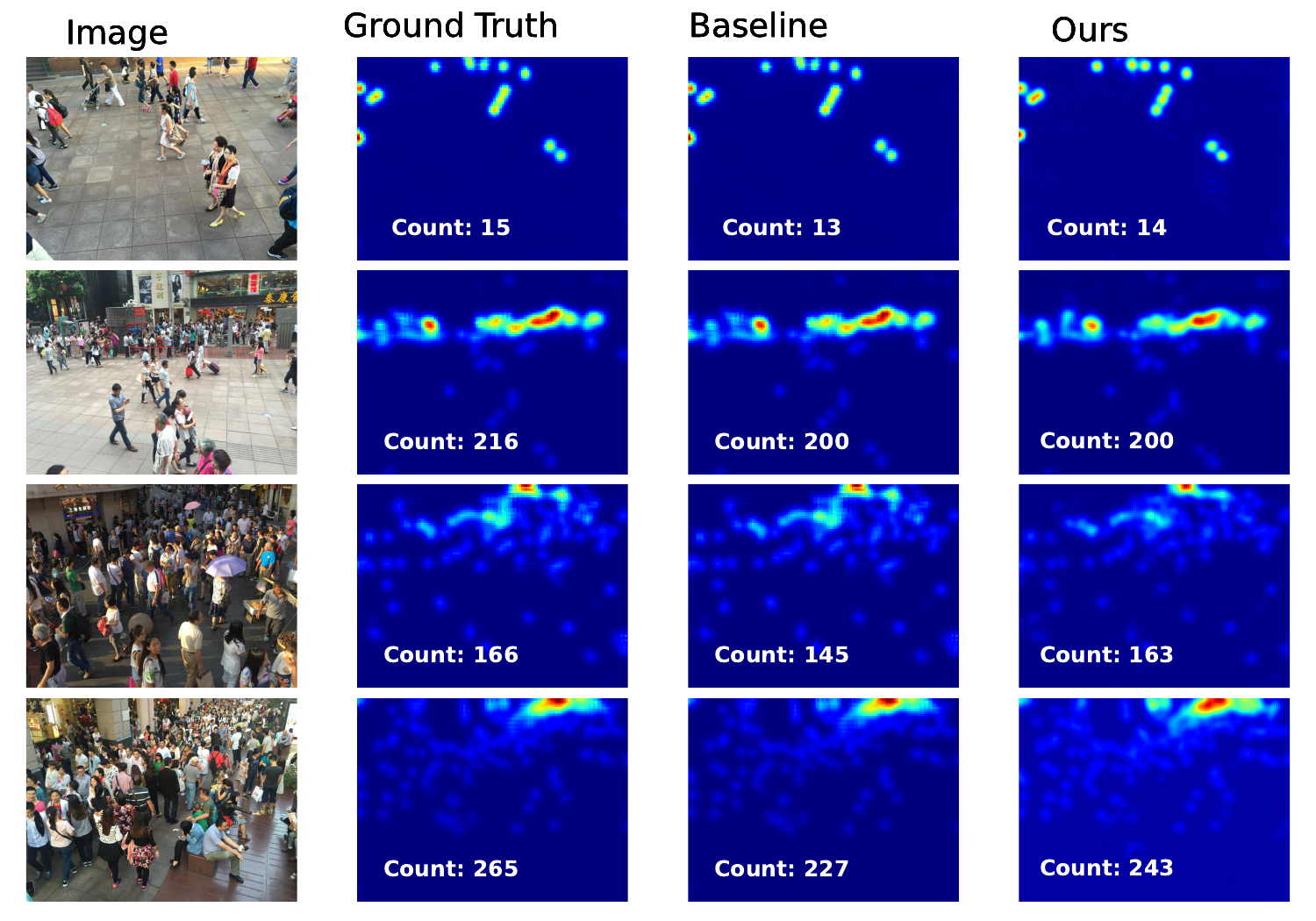}
    \caption{Visual qualitative analysis and comparison of the proposed method versus baseline models over ShanghaiTech Part B.}
    \label{fig:stb_predictions}
\end{figure}

We further evaluate the performance of \textit{TFTB} over seven crowd density estimation models. The crowd models were first trained using the standard training method using random sampling for a fixed number of epochs. Then all the models are trained using \textit{TFTB} with $\alpha={0.2, 0.3}$ for ShanghaiTech Part B and CARPK datasets, respectively. The performance was compared using the MAE and MSE metrics against all baselines over the two datasets. The results are presented in Table \ref{tab_counting}.
The results show that \textit{TFTB} consistently outperforms the baseline models when trained for an equal number of iterations. Some sample predictions generated using different baseline models and \textit{TFTB} are compared in Fig. \ref{fig:stb_predictions}.

\section{Conclusions}
In this paper, we addressed the problem of training a deep learning model under a fixed training time constraint and proposed the \textit{Train with Fixed Time Budget (TFTB)} algorithm. TFTB is an efficient method that dynamically selects important samples from a large dataset and trains a deep learning model using the reduced subset. TFTB not only benefits from the small size of the dataset to complete iterations faster than on the full dataset but also incorporates the sample importance to improve learning and convergence. Unlike traditional approaches using a fixed subset determined at the beginning of the training, TFTB dynamically ranks samples and subset the dataset at each training iteration. This approach helps in avoiding overfitting and enhances the generalization capabilities of the model. The paper presents an extensive set of 
experiments to study the performance of TFTB in the computer vision domain on both classification and regression tasks. Over 13 different models and four benchmarking datasets, our results consistently show the benefit of TFTB. We believe that TFTB can be applied in several practical applications with limited training time constraints or rapid prototyping scenarios.
\red{
A limitation of the TFTB is the use of loss-based scoring approach which is tightly coupled with the specific architecture and training dynamics of the model. For example, models with different capacities or architectures may produce varying loss distributions, leading to inconsistencies in how samples are ranked across different models. Some model-agnostic approaches can be gradient-based importance scoring and uncertainty estimates. However, these methods have their own limitations too. Another limitation of loss-based scoring method is that it focuses too much on hard samples can sometimes lead to overfitting. However, the proposed method overcomes this limitation by using the $\alpha$ parameter to control sample selection.}
In the future study, we plan to incorporate model-agnostic methods of computing sample importance in TFTB.

\section*{Acknowledgments}

\section*{Data Availability}
The ShanghaiTech dataset is publicly available in the Github repository: \newline https://github.com/desenzhou/ShanghaiTechDataset. \newline
The CARPK dataset is publicly available at: \newline https://lafi.github.io/LPN/. \newline
The CIFAR10 and CIFAR100 datasets are publicly available at: \newline https://www.cs.toronto.edu/ kriz/cifar.html.

\bibliographystyle{plainnat}
\bibliography{biblio}

\begin{thebibliography}{41}
\providecommand{\natexlab}[1]{#1}
\providecommand{\url}[1]{\texttt{#1}}
\expandafter\ifx\csname urlstyle\endcsname\relax
  \providecommand{\doi}[1]{doi: #1}\else
  \providecommand{\doi}{doi: \begingroup \urlstyle{rm}\Url}\fi

\bibitem[Bai et~al.(2020)Bai, Mao, and Chan]{Bai2020ASO}
Haoyue Bai, Jiageng Mao, and Shueng-Han~Gary Chan.
\newblock A survey on deep learning-based single image crowd counting: Network design, loss function and supervisory signal.
\newblock \emph{Neurocomputing}, 508:\penalty0 1--18, 2020.

\bibitem[Bengio et~al.(2009)Bengio, Louradour, Collobert, and Weston]{curriculum_learning_ICML2009}
Yoshua Bengio, J\'{e}r\^{o}me Louradour, Ronan Collobert, and Jason Weston.
\newblock Curriculum learning.
\newblock In \emph{Proceedings of the 26th Annual International Conference on Machine Learning}, ICML '09, page 41–48, New York, NY, USA, 2009. Association for Computing Machinery.
\newblock ISBN 9781605585161.
\newblock \doi{10.1145/1553374.1553380}.
\newblock URL \url{https://doi.org/10.1145/1553374.1553380}.

\bibitem[Casalicchio et~al.(2018)Casalicchio, Molnar, and Bischl]{Casalicchio2018}
Giuseppe Casalicchio, Christoph Molnar, and B.~Bischl.
\newblock Visualizing the feature importance for black box models.
\newblock In \emph{ECML/PKDD}, 2018.

\bibitem[Covert et~al.(2020)Covert, Lundberg, and Lee]{Covert2020}
Ian~C. Covert, Scott Lundberg, and Su-In Lee.
\newblock Understanding global feature contributions with additive importance measures.
\newblock NIPS'20, Red Hook, NY, USA, 2020. Curran Associates Inc.
\newblock ISBN 9781713829546.

\bibitem[Fan et~al.(2021)Fan, Zhang, Zhang, Lu, Zhang, and Wang]{Fan2021ASO}
Zizhu Fan, Hong Zhang, Zheng Zhang, Guangming Lu, Yudong Zhang, and Yaowei Wang.
\newblock A survey of crowd counting and density estimation based on convolutional neural network.
\newblock \emph{Neurocomputing}, 472:\penalty0 224--251, 2021.

\bibitem[Freitas et~al.(2023)Freitas, Laber, Lazera, and Molinaro]{TCL_2023}
Sergio Freitas, Eduardo Laber, Pedro Lazera, and Marco Molinaro.
\newblock Time-constrained learning.
\newblock \emph{Pattern Recognition}, 142:\penalty0 109672, 2023.
\newblock ISSN 0031-3203.

\bibitem[Friedman(2001)]{Friedman2001}
Jerome~H. Friedman.
\newblock Greedy function approximation: A gradient boosting machine.
\newblock \emph{Annals of Statistics}, 29:\penalty0 1189--1232, 2001.

\bibitem[Gao et~al.(2019)Gao, Wang, and Gao]{MobileCount_2019}
Chenyu Gao, Peng Wang, and Ye~Gao.
\newblock Mobilecount: An efficient encoder-decoder framework for real-time crowd counting.
\newblock In \emph{Pattern Recognition and Computer Vision - Second Chinese Conference, {PRCV} 2019, Xi'an, China, November 8-11, 2019, Proceedings, Part {II}}, volume 11858 of \emph{Lecture Notes in Computer Science}, pages 582--595. Springer, 2019.

\bibitem[Gao et~al.(2020)Gao, Gao, Liu, Wang, and Wang]{Gao2020CNNbasedDE}
Guangshuai Gao, Junyu Gao, Qingjie Liu, Qi~Wang, and Yunhong Wang.
\newblock Cnn-based density estimation and crowd counting: A survey.
\newblock \emph{ArXiv}, abs/2003.12783, 2020.

\bibitem[Goldstein et~al.(2013)Goldstein, Kapelner, Bleich, and Pitkin]{Goldstein2013}
A.~Goldstein, Adam Kapelner, Justin Bleich, and Emily Pitkin.
\newblock Peeking inside the black box: Visualizing statistical learning with plots of individual conditional expectation.
\newblock \emph{Journal of Computational and Graphical Statistics}, 24:\penalty0 44 -- 65, 2013.

\bibitem[Guo et~al.(2018)Guo, Huang, Zhang, Zhuang, Dong, Scott, and Huang]{Guo_2018}
Sheng Guo, Weilin Huang, Haozhi Zhang, Chenfan Zhuang, Dengke Dong, Matthew~R. Scott, and Dinglong Huang.
\newblock Curriculumnet: Weakly supervised learning from large-scale web images.
\newblock \emph{ArXiv}, abs/1808.01097, 2018.

\bibitem[Hacohen and Weinshall(2019)]{Guy_2019}
Guy Hacohen and Daphna Weinshall.
\newblock On the power of curriculum learning in training deep networks.
\newblock \emph{ArXiv}, abs/1904.03626, 2019.

\bibitem[He et~al.(2016)He, Zhang, Ren, and Sun]{ResNet_2016}
K.~He, X.~Zhang, S.~Ren, and J.~Sun.
\newblock Deep residual learning for image recognition.
\newblock In \emph{2016 IEEE Conference on Computer Vision and Pattern Recognition (CVPR)}, pages 770--778, Los Alamitos, CA, USA, jun 2016. IEEE Computer Society.

\bibitem[Hsieh et~al.(2017)Hsieh, Lin, and Hsu]{CARPK_dataset}
Meng-Ru Hsieh, Yen-Liang Lin, and Winston~H. Hsu.
\newblock Drone-based object counting by spatially regularized regional proposal network.
\newblock \emph{2017 IEEE International Conference on Computer Vision (ICCV)}, pages 4165--4173, 2017.

\bibitem[Iandola et~al.(2016)Iandola, Moskewicz, Ashraf, Han, Dally, and Keutzer]{SqueezeNet_ICLR2017}
Forrest~N. Iandola, Matthew~W. Moskewicz, Khalid Ashraf, Song Han, William~J. Dally, and Kurt Keutzer.
\newblock Squeezenet: Alexnet-level accuracy with 50x fewer parameters and <1mb model size.
\newblock \emph{ArXiv}, abs/1602.07360, 2016.

\bibitem[Jiang et~al.(2020)Jiang, Zhang, Xu, Zhang, Lv, Zhou, Yang, and Pang]{ASNet_CVPR2020}
Xiaoheng Jiang, Li~Zhang, Mingliang Xu, Tianzhu Zhang, Pei Lv, Bing Zhou, Xin Yang, and Yanwei Pang.
\newblock Attention scaling for crowd counting.
\newblock \emph{2020 IEEE/CVF Conference on Computer Vision and Pattern Recognition (CVPR)}, pages 4705--4714, 2020.

\bibitem[Jiang et~al.(2019)Jiang, Xiao, Zhang, Zhen, Cao, Doermann, and Shao]{TEDnet_CVPR2019}
Xiaolong Jiang, Zehao Xiao, Baochang Zhang, Xiantong Zhen, Xianbin Cao, David~S. Doermann, and Ling Shao.
\newblock Crowd counting and density estimation by trellis encoder-decoder networks.
\newblock \emph{2019 IEEE/CVF Conference on Computer Vision and Pattern Recognition (CVPR)}, pages 6126--6135, 2019.

\bibitem[Katharopoulos and Fleuret(2018)]{Katharopoulos2018}
Angelos Katharopoulos and François Fleuret.
\newblock Not all samples are created equal: Deep learning with importance sampling.
\newblock In \emph{International Conference on Machine Learning}, 2018.
\newblock URL \url{https://api.semanticscholar.org/CorpusID:3663876}.

\bibitem[Khan et~al.(2022)Khan, Menouar, and Hamila]{Khan2022RevisitingCC}
Muhammad~Asif Khan, Hamid Menouar, and Ridha Hamila.
\newblock Revisiting crowd counting: State-of-the-art, trends, and future perspectives.
\newblock \emph{ArXiv}, abs/2209.07271, 2022.

\bibitem[Khan et~al.(2023{\natexlab{a}})Khan, Hamila, Erbad, and Gabbouj]{DistributedIoT_Khan}
Muhammad~Asif Khan, Ridha Hamila, Aiman Erbad, and Moncef Gabbouj.
\newblock Distributed inference in resource-constrained iot for real-time video surveillance.
\newblock \emph{IEEE Systems Journal}, 17\penalty0 (1):\penalty0 1512--1523, 2023{\natexlab{a}}.
\newblock \doi{10.1109/JSYST.2022.3198711}.

\bibitem[Khan et~al.(2023{\natexlab{b}})Khan, Hamila, and Menouar]{clip_khan2023}
Muhammad~Asif Khan, Ridha Hamila, and Hamid Menouar.
\newblock Clip: Train faster with less data.
\newblock In \emph{2023 IEEE International Conference on Big Data and Smart Computing (BigComp)}, pages 34--39. IEEE, 2023{\natexlab{b}}.

\bibitem[Khan et~al.(2023{\natexlab{c}})Khan, Menouar, and Hamila]{DroneNet2023}
Muhammad~Asif Khan, Hamid Menouar, and Ridha Hamila.
\newblock Dronenet: Crowd density estimation using self-onns for drones.
\newblock In \emph{2023 IEEE 20th Consumer Communications \& Networking Conference (CCNC)}, pages 455--460, 2023{\natexlab{c}}.
\newblock \doi{10.1109/CCNC51644.2023.10059904}.

\bibitem[Khan et~al.(2023{\natexlab{d}})Khan, Menouar, and Hamila]{LCDnet_khan2023}
Muhammad~Asif Khan, Hamid Menouar, and Ridha Hamila.
\newblock Lcdnet: A lightweight crowd density estimation model for real-time video surveillance.
\newblock \emph{J. Real-Time Image Process.}, 20\penalty0 (2), mar 2023{\natexlab{d}}.
\newblock ISSN 1861-8200.
\newblock \doi{10.1007/s11554-023-01286-8}.
\newblock URL \url{https://doi.org/10.1007/s11554-023-01286-8}.

\bibitem[Khan et~al.(2024)Khan, Menouar, and Hamila]{CLCC_2024}
Muhammad~Asif Khan, Hamid Menouar, and Ridha Hamila.
\newblock Curriculum for crowd counting--is it worthy?
\newblock \emph{arXiv preprint arXiv:2401.07586}, 2024.

\bibitem[Li et~al.(2019)Li, Yumer, and Ramanan]{li2019budgeted}
Mengtian Li, Ersin Yumer, and Deva Ramanan.
\newblock Budgeted training: Rethinking deep neural network training under resource constraints.
\newblock In \emph{International Conference on Learning Representations}, 2019.

\bibitem[Li et~al.(2021)Li, Cao, Wang, Chen, and Feng]{Li_2021}
Wenxi Li, Zhuoqun Cao, Qian Wang, Songjian Chen, and Rui Feng.
\newblock Learning error-driven curriculum for crowd counting.
\newblock \emph{2020 25th International Conference on Pattern Recognition (ICPR)}, pages 843--849, 2021.

\bibitem[Li et~al.(2018)Li, Zhang, and Chen]{CSRNet_CVPR2018}
Yuhong Li, Xiaofan Zhang, and Deming Chen.
\newblock Csrnet: Dilated convolutional neural networks for understanding the highly congested scenes.
\newblock \emph{2018 IEEE/CVF Conference on Computer Vision and Pattern Recognition}, pages 1091--1100, 2018.

\bibitem[Lundberg et~al.(2018)Lundberg, Erion, and Lee]{Lundberg2018}
Scott~M. Lundberg, Gabriel~G. Erion, and Su-In Lee.
\newblock Consistent individualized feature attribution for tree ensembles.
\newblock \emph{ArXiv}, abs/1802.03888, 2018.

\bibitem[Paul et~al.(2021)Paul, Ganguli, and Dziugaite]{Paul2021}
Mansheej Paul, Surya Ganguli, and Gintare~Karolina Dziugaite.
\newblock Deep learning on a data diet: Finding important examples early in training.
\newblock In \emph{NeurIPS}, 2021.

\bibitem[Poulinakis et~al.(2023)Poulinakis, Drikakis, Kokkinakis, and Spottswood]{ref1}
Konstantinos Poulinakis, Dimitris Drikakis, Ioannis~W. Kokkinakis, and Stephen~Michael Spottswood.
\newblock Machine-learning methods on noisy and sparse data.
\newblock \emph{Mathematics}, 11\penalty0 (1), 2023.
\newblock ISSN 2227-7390.
\newblock \doi{10.3390/math11010236}.
\newblock URL \url{https://www.mdpi.com/2227-7390/11/1/236}.

\bibitem[Sandler et~al.(2018)Sandler, Howard, Zhu, Zhmoginov, and Chen]{MobileNetV2_2018}
Mark Sandler, Andrew~G. Howard, Menglong Zhu, Andrey Zhmoginov, and Liang-Chieh Chen.
\newblock Mobilenetv2: Inverted residuals and linear bottlenecks.
\newblock \emph{2018 IEEE/CVF Conference on Computer Vision and Pattern Recognition}, pages 4510--4520, 2018.

\bibitem[Sindagi and Patel(2017)]{CMTL_AVSS2017}
Vishwanath~A. Sindagi and Vishal~M. Patel.
\newblock Cnn-based cascaded multi-task learning of high-level prior and density estimation for crowd counting.
\newblock \emph{2017 14th IEEE International Conference on Advanced Video and Signal Based Surveillance (AVSS)}, pages 1--6, 2017.

\bibitem[Sofos et~al.(2024)Sofos, Drikakis, and Kokkinakis]{ref2}
Filippos Sofos, Dimitris Drikakis, and Ioannis~William Kokkinakis.
\newblock {Deep learning architecture for sparse and noisy turbulent flow data}.
\newblock \emph{Physics of Fluids}, 36\penalty0 (3):\penalty0 035155, 03 2024.
\newblock ISSN 1070-6631.
\newblock \doi{10.1063/5.0200167}.
\newblock URL \url{https://doi.org/10.1063/5.0200167}.

\bibitem[Song et~al.(2021)Song, Wang, Wang, Tai, Wang, Li, Wu, and Ma]{SASNet_AAAI2021}
Qingyu Song, Changan Wang, Yabiao Wang, Ying Tai, Chengjie Wang, Jilin Li, Jian Wu, and Jiayi Ma.
\newblock To choose or to fuse? scale selection for crowd counting.
\newblock In \emph{AAAI}, 2021.

\bibitem[Szegedy et~al.(2015)Szegedy, Liu, Jia, Sermanet, Reed, Anguelov, Erhan, Vanhoucke, and Rabinovich]{GoogLeNet_2015}
Christian Szegedy, Wei Liu, Yangqing Jia, Pierre Sermanet, Scott Reed, Dragomir Anguelov, Dumitru Erhan, Vincent Vanhoucke, and Andrew Rabinovich.
\newblock Going deeper with convolutions.
\newblock In \emph{2015 IEEE Conference on Computer Vision and Pattern Recognition (CVPR)}, pages 1--9, 2015.
\newblock \doi{10.1109/CVPR.2015.7298594}.

\bibitem[Yang and Zhu(2022)]{Yang2022SurveyOA}
Ge~Yang and Dian Zhu.
\newblock Survey on algorithms of people counting in dense crowd and crowd density estimation.
\newblock \emph{Multimedia Tools and Applications}, 82:\penalty0 13637--13648, 2022.

\bibitem[Yang et~al.(2022)Yang, Xie, Peng, Xu, Sun, and Li]{Yang2022}
Shuo Yang, Zeke Xie, Hanyu Peng, Minjing Xu, Mingming Sun, and P.~Li.
\newblock Dataset pruning: Reducing training data by examining generalization influence.
\newblock \emph{ArXiv}, abs/2205.09329, 2022.

\bibitem[Zayed et~al.(2022)Zayed, Parthasarathi, Mordido, Palangi, Shabanian, and Chandar]{Zayed2022}
Abdelrahman Zayed, Prasanna Parthasarathi, Gonçalo Mordido, Hamid Palangi, Samira Shabanian, and Sarath Chandar.
\newblock Deep learning on a healthy data diet: Finding important examples for fairness.
\newblock \emph{ArXiv}, abs/2211.11109, 2022.

\bibitem[Zhang et~al.(2016)Zhang, Zhou, Chen, Gao, and Ma]{MCNN_CVPR2016}
Yingying Zhang, Desen Zhou, Siqin Chen, Shenghua Gao, and Yi~Ma.
\newblock Single-image crowd counting via multi-column convolutional neural network.
\newblock In \emph{2016 IEEE Conference on Computer Vision and Pattern Recognition (CVPR)}, pages 589--597, 2016.
\newblock \doi{10.1109/CVPR.2016.70}.

\bibitem[Zhou et~al.(2023)Zhou, Yang, Ji, and Zhu]{reviewer1b}
Z~Zhou, X~Yang, H~Ji, and Z~Zhu.
\newblock {Improving the classification accuracy of fishes and invertebrates using residual convolutional neural networks}.
\newblock \emph{ICES Journal of Marine Science}, 80\penalty0 (5):\penalty0 1256--1266, 04 2023.
\newblock ISSN 1054-3139.
\newblock \doi{10.1093/icesjms/fsad041}.
\newblock URL \url{https://doi.org/10.1093/icesjms/fsad041}.

\bibitem[Zhou et~al.(2024)Zhou, Hu, Yang, and Yang]{reviewer1a}
Zhiyu Zhou, Yanjun Hu, Xingfan Yang, and Junyi Yang.
\newblock Yolo-based marine organism detection using two-terminal attention mechanism and difficult-sample resampling.
\newblock \emph{Applied Soft Computing}, 153:\penalty0 111291, 2024.
\newblock ISSN 1568-4946.
\newblock \doi{https://doi.org/10.1016/j.asoc.2024.111291}.

\end{thebibliography}

\end{document}